\documentclass[10pt,twocolumn,letterpaper]{article}

\usepackage[pagenumbers]{cvpr} %
\usepackage{pifont}
\usepackage{graphicx}
\usepackage{amsmath}
\usepackage{amssymb}
\usepackage{booktabs}

\usepackage[title]{appendix}
\usepackage{color, soul}
\usepackage{nicematrix}
\usepackage{multirow}
\usepackage[inline]{enumitem}
\usepackage{relsize}

\usepackage{minorrevision}

\def\taskname{commented retrieval\xspace}
\def\tasknamebos{Commented retrieval\xspace}
\def\tasknametitle{Commented Retrieval\xspace}
\def\tasknameshort{CoR\xspace}
\def\modelname{UniCoRN\xspace}

\def\cirr{CIRR\xspace}
\def\fashioniq{Fashion-IQ\xspace}
\def\wikicomment{Wiki-\tasknameshort}
\def\wikimm{WikiWeb2M\xspace}
\def\cirrcomment{CIRR-\tasknameshort}
\def\oven{OVEN\xspace}
\def\infoseek{InfoSeek\xspace}

\newcommand{\textmod}{\ensuremath{{\langle t\rangle}}}
\newcommand{\imgmod}{\ensuremath{{\langle i\rangle}}}
\newcommand{\retrmod}{\ensuremath{{\langle r\rangle}}}
\newcommand{\mmmod}{\ensuremath{\text{MM}}}
\newcommand{\lmmmod}{\ensuremath{\text{LMM}}}

\DeclareMathOperator*{\argmax}{argmax}

\newcommand{\retrieverparams}{\ensuremath{\theta}\xspace}
\newcommand{\retriever}{\ensuremath{R_\retrieverparams}\xspace}
\newcommand{\hiddenstate}{\ensuremath{H_{\text{LMM}}}\xspace}

\definecolor{cvprblue}{rgb}{0.21,0.49,0.74}
\usepackage[pagebackref,breaklinks,colorlinks,allcolors=cvprblue]{hyperref}

\def\paperID{10034} %
\def\confName{CVPR}
\def\confYear{2025}

\title{\modelname: Unified \tasknametitle Network with LMMs}

\author{
{Maximilian Jaritz \quad Matthieu Guillaumin \quad Sabine Sternig \quad Loris Bazzani}
\\
Amazon.com
}

\begin{document}
\maketitle
\begin{abstract}
Multimodal retrieval methods have limitations 
in handling complex, compositional queries that require reasoning about the 
visual content of both the query and the retrieved entities.
On the other hand, Large Multimodal Models (LMMs) can answer with language to more complex visual questions, but without the inherent ability to retrieve relevant entities to support their answers.
We aim to address these limitations with \modelname, a Unified Commented Retrieval Network that combines the strengths of composed multimodal retrieval methods and generative language approaches, going beyond Retrieval-Augmented Generation~(RAG). %
We introduce an entity adapter module to inject the retrieved multimodal entities back into the LMM, so it can attend to them while generating answers and comments.
By keeping the base LMM frozen, \modelname preserves its original capabilities while being able to perform both retrieval and text generation tasks under a single integrated framework.
To assess these new abilities, we introduce the Commented Retrieval task (\tasknameshort) and a corresponding dataset, with the goal of retrieving an image that accurately answers a given question and generate an additional textual response that provides further clarification and details about the visual information.
We demonstrate the effectiveness of \modelname on several datasets 
showing improvements of +4.5\% recall over the state of the art for composed multimodal retrieval and of 
+14.9\% METEOR / +18.4\% BEM over RAG for commenting in \tasknameshort.
\end{abstract}
\section{Introduction}
\label{sec:introduction}
Multimodal retrieval, the task of retrieving relevant information across text, visual and other modalities, has gained significant attention in recent years~\cite{radford2021clip,li2022blip,zhai2023sigmoid,chen2023pali, yu2022coca, wang2023image, zhou2022non, zhai2019large, yao2021filip, goel2022cyclip, gao2022pyramidclip, lee2022uniclip, ilharco2021openclip, schuhmann2021laion400m, gao2023softclip, bao2022vlmo, desai2023hyperbolic} demonstrating strong performance on a variety of tasks in supervised, zero-shot, and out-of-domain applications.
However, recent methods still struggle to capture complex, compositional requests that require specific reasoning about visual content.
On the other hand, LMMs~\cite{openai2024gpt4,geminiteam2024gemini,anthropic2024claude,zhu2024minigpt,liu2023llava,liu2023improvedllava,chen2024far,jiang2024mantis, wang2024qwen2, ye2024mplug, chen2024internvl, mckinzie2024mm1, wu2023next, dubey2024llama, agrawal2024pixtral} can deal with complex visio-linguistic requests with language output, without the inherent ability to retrieve and output relevant images to support the answer.
This raises an intriguing question for which we seek to make advancements in our work: \emph{How can we integrate the multimodal reasoning capabilities of LMMs with the retrieval capabilities of composed multimodal models?} 

\begin{figure}[t]
  \centering
   \includegraphics[width=1\linewidth]{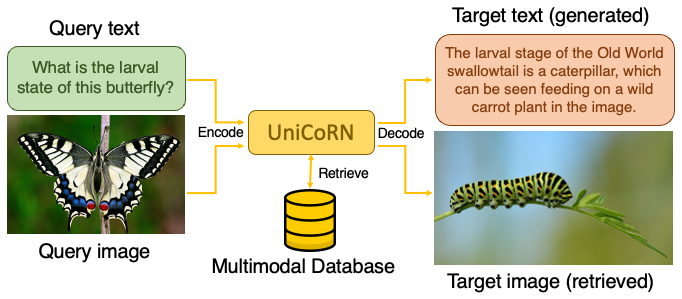}
   \vspace{-2mm}
   \caption{\underline{\tasknamebos}. Given an query image and question, \modelname can retrieve an image from a database and can produce a textual answer that offers further clarification and details.}
   \label{fig:main_figure}
   \vspace{-4mm}
\end{figure}

Let us consider a user asking ``what is the larval state of this butterfly?'' as in Fig.~\ref{fig:main_figure}.
The model must not only recognize the species of butterfly, but also understand the query and find an image depicting the corresponding caterpillar stage, which has little visual similarity with the query.
While prior work~\cite{han2017automatic,ak2018learning,Vo_2019_CVPR, guo2018dialog,guo2019fashion, hou2021learning,chen2020image, Baldrati_2022_CVPR, Goenka_2022_CVPR} has explored composed retrieval techniques for such structured requests, these approaches have limitations in providing comprehensive responses.
On the other hand, Visual Question Answering (VQA) methods~\cite{antol2015vqa, wu2017visual, anderson2018bottom, schwenk2022okvqa, singh2019towards, shah2019kvqa, goyal2017making, gurari2018vizwiz, yan-xie-2024-echosight}, even when leveraging LMMs, are restricted to answering solely with language.
An ideal model should both retrieve a relevant image and generate explanatory text that complements and further explains the visual information.
Referring back to Fig.~\ref{fig:main_figure}, the model returns an image of the butterfly's larval stage and a comment, \eg ``The larval stage of the Old World swallowtail is a caterpillar, which can be seen feeding on a wild carrot plant in the image.''.
We refer to this novel multimodal task as \emph{\tasknametitle}~(\tasknameshort).

In this work, we propose \modelname, a novel Unified Commented Retrieval Network that leverages the complementary strengths of discriminative cross-modal matching and generative language models to perform \tasknameshort. 
First, we present a retrieval module that aligns the hidden state of LMMs with the retrieval space of CLIP models.
This allows us to integrate the deep visio-linguistic reasoning of LMMs with powerful multimodal and cross-modal retrieval capabilities. 
The retrieval model is trained to be aware of the complex answers and comments, rather than simple image captions like in previous work~\cite{wei2023uniir, lin2024mmembeduniversalmultimodalretrieval, jiang2024vlm2vec}.
Second, we introduce an entity adapter module to inject the retrieved entities back into the LMM, so they can condition the generated answers and comments.
This technique enables interleaved, integrated multimodal output at inference time, which goes beyond post-processing of output tokens
or prompt engineering of the LMM, \eg with RAG~\cite{lewis2021retrievalaugmentedgenerationknowledgeintensivenlp}.
By training these new modules for \tasknameshort, we strengthen the connection between the initial question and the retrieved entities, enabling the generation of more coherent and relevant textual responses.
Notably, our contributions are additions to a frozen LMM, so, unlike~\cite{jiang2024vlm2vec, lin2024mmembeduniversalmultimodalretrieval}, we guarantee the preservation of all its original capabilities (such as captioning, VQA, grounding, and more) within our unified framework.

Existing multimodal tasks and datasets~\cite{hu2023open, kazemzadeh2014referitgame, chen2023can, wu2021fashion, liu2021image} have simplistic (1-3 words) answers that do not explain why and how the retrieved entity is relevant to the initial query, and thus are not well suited for training and evaluating models for  \tasknameshort.
In this work, we introduce two challenging human-curated \tasknameshort datasets based on the \cirr~\cite{Liu_2021_ICCV} and \wikimm dataset~\cite{burns2023wiki}, that blend aspects of composed retrieval and VQA. 
Given an input image and question, the goal is to retrieve from a large candidate pool the entity that answers the question, and produce an additional textual response that offers further clarification and details (Fig.~\ref{fig:main_figure}). 
This task echoes research showing that humans learn better when they can jointly examine textual responses alongside the relevant visuals~\cite{mayer2003three}.
Our evaluations on diverse datasets for composed retrieval and \taskname show significant improvements of \modelname over state-of-the-art models.
In particular, we show an average improvement for composed retrieval of $+4.5\%$ on \fashioniq, \cirr, \oven, \infoseek, and the proposed \wikicomment dataset in terms of recall when compared to UniIR~\cite{wei2023uniir}.
Moreover, we show an improvement of +14.9\% in terms of METEOR score over a RAG approach combining UniIR and InternVL2 on the \tasknameshort task.
\section{Related Work}
\label{sec:related_work}

\noindent\textbf{Instructable Retrieval.}
Multimodal representation for retrieval~\cite{radford2021clip, li2022blip, zhai2023sigmoid, chen2023pali, yu2022coca, wang2023image, zhou2022non, zhai2019large, yao2021filip, gao2022pyramidclip, lee2022uniclip, ilharco2021openclip, schuhmann2021laion400m, bao2022vlmo} has been a prominent area of research demonstrating strong performance. %
To handle compositional requests that require reasoning about the visual content, recent work focused on composed retrieval~\cite{vo2019composing, Baldrati_2022_CVPR, Baldrati_2023_ICCV, Goenka_2022_CVPR, Chen_2020_CVPR, Wu_2021_CVPR, Gu_2024_CVPR, Vaze_2023_CVPR, Suo_2024_CVPR, Wan_2024_CVPR, Liu_2024_WACV, karthik2023vision}.
However, these methods focus on limited specific retrieval domains and can only follow predefined instruction templates.
To address this limitation, recent models have been made instructable~\cite{zhang2024magiclens, wei2023uniir, jiang2024vlm2vec, lin2024mmembeduniversalmultimodalretrieval, qi2024roravlm, karthik2023vision, levy2024chatting} to capture richer multimodal relationships. 
MagicLens~\cite{zhang2024magiclens} is a simple instructable model trained on a curated large dataset including instructions, while UniIR~\cite{wei2023uniir} proposed training CLIP/BLIP variants conditioned on prompted instructions.
While these approaches have demonstrated significant progress, they lack expressiveness in the multimodal output space due to the limitations of CLIP models. 
In contrast, methods based on LMMs~\cite{lin2024mmembeduniversalmultimodalretrieval, jiang2024vlm2vec} that are fine-tuned for retrieval do not preserve the original capabilities of the LMMs.
\modelname aims to overcome this issue and preserve capabilities by design by enabling a frozen LMM to retrieve relevant content and generate textual responses tailored to both the input question and the retrieved visual content.

\noindent\textbf{Retrieval via Generation.}
LMMs~\cite{openai2024gpt4,geminiteam2024gemini,anthropic2024claude,zhu2024minigpt,liu2023llava,liu2023improvedllava,chen2024far,jiang2024mantis, wang2024qwen2, ye2024mplug, chen2024internvl, mckinzie2024mm1, wu2023next, dubey2024llama, agrawal2024pixtral, lin2024mmembeduniversalmultimodalretrieval} have demonstrated remarkable reasoning capabilities when processing language and visual elements (see survey in~\cite{yin2023survey}).
Recently, LMMs were combined with diffusion models to generate images along with textual answers~\cite{tian2024mminterleaved, li2023textbind, koh2023grounding, wu2023next, ge2023making, yu2023language, team2024chameleon}.
Despite providing an interesting user experience, the generated images are by nature not grounded to real entities, which is critical for retrieval applications, such as online shopping, news, or Wikipedia.
These models are based on memory-intensive diffusion models and trained with a reconstruction objective that is not in line with retrieval.
In contrast, \modelname uses a retrieval approach that preserves the factuality of the images that are shown to the user, while commenting them in a generative way. 
It is lightweight, as it does not use diffusion models, and the LMMs are frozen, thus preserving their reasoning capabilities of other non-retrieval tasks.

\noindent\textbf{Retrieval-Augmented Generation (RAG).}
The parametric way that language models are trained and store knowledge limits the ease to expand or revise their memory with more recent information.
RAG~\cite{lee2019latent, lewis2020retrieval, guu2020retrieval, karpukhin2020dense,fevry2020entities} and Multimodal RAG~\cite{qi2024roravlm, hu2023reveal, yu2024visrag, chen2022murag, qu2024alleviating, caffagni2024wiki, yan-xie-2024-echosight} have been proposed as a non-parametric way to provide up-to-date knowledge to an LMM by adding relevant content to the prompt before processing by the LMM.
Using this augmented prompt, the LMM can select, summarize and alter the entities to generate its output.
\modelname differs from traditional RAG in that it processes user input via the LMM to build a better query and outputs the retrieved entity unaltered.
Then, \modelname's entity adapter module is trained to feed optimized entity representations to the LMM, which guides the extraction of the information necessary to answer the request.

\noindent\textbf{\tasknametitle (\tasknameshort).}
\tasknameshort is an underexplored task that requires not only retrieving relevant images, as in composed retrieval, but also generating plausible textual answers that refer to the image and contain additional information.
While preliminary work~\cite{wei2023uniir, jiang2024vlm2vec, qi2024roravlm, karthik2023vision} and datasets~\cite{hu2023open, chen2023can, mensink2023encyclopedic} focuse on some aspects of this task, they have two key limitations: 
1) The complexity of the textual answers is constrained, typically limited to a few words or basic captions.
2) The textual answers are aligned with the image, either providing redundant information (e.g., image captions) and failing to add useful complementary information.
To fully cover the \tasknameshort task, we claim that the answers should be multimodal in nature, with the image and text working in tandem to provide a richer, more comprehensive response.
The modalities should have a clear connection, yet offer complementary information that enhances the overall response.
To alleviate the limitations of existing \tasknameshort datasets, we created two challenging human-curated ones that we hope will foster future research in this area.
\section{\modelname}
\label{sec:method}

With \modelname, we propose to expand the capabilities of a frozen base LMMs with two novel interconnected modules: 
1)~Comment-aware retrieval, which is responsible for understanding the visio-linguistic representation of the query question and image, and projecting it to the target comment and image space for retrieval (Sec.~\ref{ssec:comment_retrieval}); 
2)~Retrieval-aware generation, which learns to leverage the retrieved entity to generate an answer to the user question (Sec.~\ref{ssec:retrieve_generate}).

\subsection{Comment-aware Retrieval}
\label{ssec:comment_retrieval}

\begin{figure}[t]
  \centering
   \includegraphics[width=.8\linewidth]{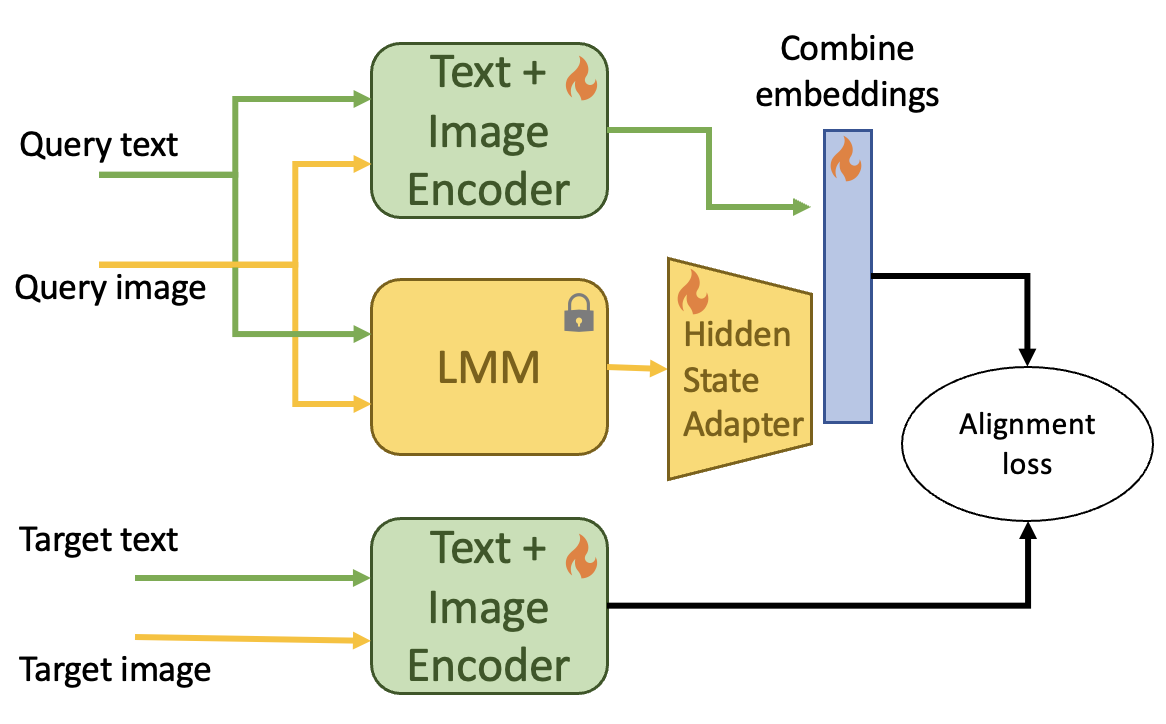}
   \vspace{-3mm}
   \caption{\underline{Comment-aware Retrieval}. The query is inputed to both a CLIP-trained image-text encoder and an LMM. The LMM representation is projected to the space of the image-text encoder. Alignment of query and targets is done using contrastive loss.}
   \label{fig:comment_retrieval}
   \vspace{-3mm}
\end{figure}

Given a query text and image $q = (q^\textmod, q^\imgmod)$, we aim to train a retrieval model \retriever, with parameters \retrieverparams, that is capable of understanding $q$ and retrieve a related target entity composed of text and image $d = (d^\textmod, d^\imgmod)$, such that:
\begin{equation}
    d = \argmax_{d' \in \mathcal{D}}~{\retriever(q, d')} \label{eq:retrieval}
\end{equation}
where $\mathcal{D}$ is a database of multimodal entities (a.k.a documents).
The best definition for the retriever \retriever depends on the target application.
For cross-modal and multimodal retrieval, CLIP-like models~\cite{radford2021clip, gao2023softclip, gao2022pyramidclip, ilharco2021openclip, wei2023uniir} are the \emph{de-facto} choice based on their 
performance in diverse scenarios. %
However, they are limited in the visio-linguistic understanding capabilities, especially when moving to more complex applications like composed retrieval and \taskname.
To address this limitation, we leverage the reasoning capabilities of LMMs in the training process of the retriever integrated with CLIP, as shown in Fig.~\ref{fig:comment_retrieval}.

The multimodal query $q$ is fed into the base LMM to obtain its output hidden state  $h\!=\!\hiddenstate(q)$.
In parallel, the query is fed into a multimodal encoder $\psi_\mmmod(q) = \psi^\textmod(q^\textmod) + \psi^\imgmod(q^\imgmod)$, where $\psi^\textmod$ and $\psi^\imgmod$ are text and image CLIP-like encoders, respectively.
Then, we define a Hidden State Adapter $\psi_\lmmmod$ to map $h$ into the same space as $\psi_\mmmod$, as follows:
\begin{equation}
    \psi_\lmmmod(h) = \text{FC}(\text{GeLU}(\text{FC}(h)))
\end{equation}
where FCs are fully connected layers and GeLU is the activation function.
The final embedding for $q$ combines the CLIP-based embeddings and the adapter output:
\begin{equation}
    \psi(q) = \beta \cdot \psi_\lmmmod(\hiddenstate(q)) + (1-\beta) \cdot \psi_\mmmod(q) \label{eq:mm_fusion}
\end{equation}
where $\beta$ is a learnable weight part of $\theta$.
Note, this score-fusion approach to CLIP is inspired by UniIR~\cite{wei2023uniir}, where the authors show this is one of the most effective ways to combine multimodal inputs for multimodal retrieval.

Finally, we define \retriever as the dot product of query and document embeddings: $\retriever(q, d) = \psi(q)^T\!\cdot\!\psi_\mmmod(d)$.
As shown in Fig.~\ref{fig:comment_retrieval}, the multimodal target documents are not using the LMM and therefore only use the multimodal encoder $\psi_\mmmod$.
This asymmetrical approach between query and document has several advantages: 1) it saves cost during the indexing phase by avoiding to use an expensive LMM on the potentially large pool of multimodal documents, and 2) it bridges the domain gap between the queries and the database via fine-tuning of the adapter. 
In contrast with other methods that use LMMs for indexing~\cite{jiang2024vlm2vec, lin2024mmembeduniversalmultimodalretrieval}, our method keeps indexing lightweight.

We train our retriever using contrastive loss~\cite{radford2021clip}. Unlike previous work~\cite{wei2023uniir, lin2024mmembeduniversalmultimodalretrieval,jiang2024vlm2vec}, we combine target captions and more complex target comments that cover aspects beyond the query and target images, as we will describe in Sec.~\ref{sec:datasets}.
This makes our model aware of complex comments during training and thus will better fit with retrieval-aware generation as described in Sec.~\ref{ssec:retrieve_generate}.

\begin{figure}[t]
  \centering
   \includegraphics[width=1\linewidth]{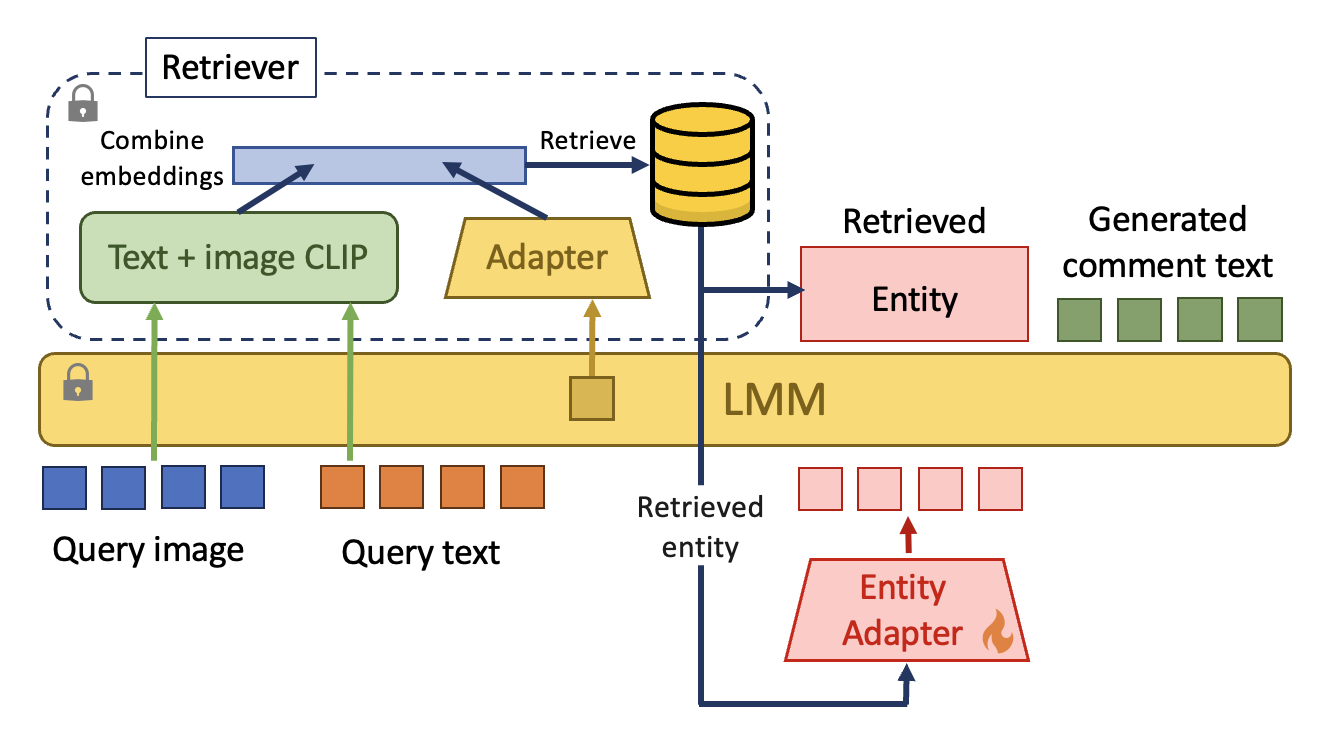}
   \vspace*{-8mm}
   \caption{\underline{Retrieval-aware Generation}. The query image and text are fed to the LMM, which asks the retriever for relevant entities.
   The best entity is provided to the user and adapted into the LMM, so it can attend to it for generating a useful comment.}
   \label{fig:retrieve_generate}
   \vspace{-3mm}
\end{figure}

\subsection{Retrieval-aware Generation}
\label{ssec:retrieve_generate}

We now detail how we enable an LMM to ingest the query question and image, encode retrieved entities, and generate an answer to the question.
The process is shown in Fig.~\ref{fig:retrieve_generate}.
\vspace*{-3mm}
\subsubsection{Retrieved Entities as Input Modality}

In essence, a Large Language Model (LLM) predicts the probability of the next token $\tau_n\!\in\!\{1, \ldots, V\}$, where $V$ is the vocabulary size, given an input sequence of $n$ tokens: $p(\tau_{n} | \tau_0, \ldots, \tau_{n-1})$.
The first layer of the LLM projects the tokens into an input embedding space, mapping each of the $V$ possible \emph{text} tokens to a vector $\varphi^\textmod(\tau_k)\!=\!\phi^\textmod_k \in \mathbb{R}^{d}$, where $d$ is the input embedding size of the LLM.
Therefore, we can also see the LLM as predicting the next token based on a sequence of text-based embeddings:
\begin{equation}
    p(\tau_{n} ~ | ~ \phi^\textmod_0, \ldots, \phi^\textmod_{n-1}).
    \label{eq:llm_embd}
\end{equation}
LMMs~\cite{liu2023improvedllava, chen2024internvl} extend this principle by adapting visual data into input image tokens $\tau^\imgmod$ or embeddings $\phi^\imgmod$.
Notably, Eq.~\ref{eq:llm_embd} is unchanged for LMMs, except that the inputs are now a mix of text $\phi^\textmod$ and image $\phi^\imgmod$ embeddings in $\mathbb{R}^d$.

As described in Sec.~\ref{ssec:comment_retrieval}, the retriever finds the most relevant multimodal document $d_m$ in the database based on the multimodal query $q$ and the projected hidden state of the LMM.
As an important contribution of our method, we then inject $d_m$ into the LMM as a new input modality $\retrmod$ in order to condition the rest of the generation process, and therefore ensure that the model attends to it while generating the textual comment.
This is important because of the finite nature of any database: the retrieved documents can substantially differ from the vector used as query.
Thus, using solely the user input to condition the generation may lead to comments that are irrelevant to the retrieved entity.

Similarly to other input modalities, features of the retrieved document are extracted and adapted into the input embedding space of the LMM:
\begin{equation}
  \varphi^\retrmod(d_m) = [\phi^\retrmod_{m,0}, \ldots, \phi^\retrmod_{m,l_m}] = \phi^\retrmod_m \in \mathbb{R}^{d\times l_m},
\end{equation}
where $l_m$ is the length of the representation of the document $d_m$.
In this way, the LMM can attend to the retrieved data in addition to all previous inputs as it continues generation:
\begin{equation}
    p(\tau_{n} ~ | \cdots, \phi^\textmod_k, \cdots, \phi^\imgmod_p, \cdots, \phi^\retrmod_m, \cdots).
    \label{eq:llm_embd2}
\end{equation}
However, the frozen LMM was not trained to distinguish entities that are retrieved from a database and treat them appropriately to generate a convincing answer.
Thus, we make the entity adapter a trainable module as described next.

\subsubsection{Entity Adapter Module}

The definition of $\varphi^\retrmod(d_m)$ depends on the nature of the dataset and the entities it contains: text, images, audio, video, pre-computed features, or any combination thereof.
We propose to design an Entity Adapter module $\varphi^\retrmod_{\xi}(d_m)$, as shown in Fig.~\ref{fig:retrieve_generate}, with parameters $\xi$, which can be fine-tuned on training data.
More specifically for our \tasknameshort tasks, the entities are images, or multimodal combinations of images and text data (e.g. titles, captions, ...), that are used to generate the answer.
As a consequence, the Entity Adapter treats the various modalities differently, \ie, $\varphi^\retrmod_{\xi}(d_m) = (\varphi^\imgmod_{\xi}(d^\imgmod_m), \varphi^\textmod_{\xi}(d^\textmod_m))$.
In our datasets, the textual components are natural text, so we provide them directly to the LMM, without fine-tuning the token embeddings, \ie, $\varphi^\textmod_{\xi}(d_m)=\phi^\textmod$.
For the visual part, we use the same architecture as the base LMM~\cite{chen2024internvl}, with the adapter:
\begin{equation}
\varphi^\imgmod_{\xi}(d_m)=\text{FC}(\text{GeLU}(\text{FC}(\text{LayerNorm}(x_m)))), \label{eq:entityadapter}
\end{equation}
where $\{x_m\}$ are the ViT image features of the variable number of tiles of the image of $d^\imgmod_m$.

\subsubsection{Training}
To train our novel Entity Adapter, we leverage a dataset $\mathcal{D}$ of ground-truth multimodal triplets $(q, d, c)$, where $q$ is the multimodal query, $d$ is the ground-truth target document, and $c$ is the ground-truth answer that relates the query and the target.
The training of $\xi$ is performed for the task of next token prediction of $c$, using the cross-entropy loss:
\begin{equation}
\mathcal{L}_{CE}(\xi) = \sum_{(q, d, c)\in\mathcal{D}} \quad
\sum_{\tau_n\in c} -\delta_n.\log(p_n(\xi)),
\end{equation}
where $p_n(\xi)\!=\!p(\tau_n | \phi^\imgmod_q, \phi^\textmod_q, \nonumber \varphi^\retrmod_{\xi}(d), \phi^\textmod_0, \ldots, \phi^\textmod_{n-1})$,
and $\delta_n$ is the ground-truth distribution for $\tau_n$.
As shown in Fig.~\ref{fig:retrieve_generate}, in practice, all other parts of the model  are kept frozen during training and only the adapter in Eq.~\ref{eq:entityadapter} is trained. 
For faster convergence, this adapter is initialized from the pretrained visual adapter of the original LMM.

With this approach, despite keeping the base LMM frozen, we are achieving two objectives: \begin{enumerate*}[label=\arabic*)]\item we align the multimodal entities with the LMM input embeddings, thus helping the LMM extract the relevant information from the entities and closing any domain gap between the data used to pre-train the LMM and the retrieval dataset, and \item we encode commenting instructions in the adapter, thus optimizing the inputs to the LMM to bias it to generate useful, high-quality comments for documents of this dataset. %
\end{enumerate*}
As a consequence, we train separate, specialized Entity Adapters for different tasks and datasets.

\section{\tasknametitle~(\tasknameshort) Task}
\label{sec:datasets}

As already discussed in related work, the related task of composed retrieval is well explored in the literature, and there exist a number of datasets for it.
Some datasets for this task, \eg, \cirr~\cite{Liu_2021_ICCV} and \fashioniq~\cite{guo2019fashion}, include multimodal queries, while the targets are only images.
Other relevant datasets, \eg, \oven~\cite{hu2023open} and \infoseek~\cite{chen2023can}, have both multimodal queries and targets, however the connection between inputs and outputs are simple high-level entities.
Overall, existing datasets either lack target text altogether or feature overly simplistic textual responses that are directly related to the target image. 
These datasets are not well-suited to evaluate the full extent of the \taskname task, where the goal is not only to retrieve a relevant target image, but also to generate a detailed, complex textual response that refer to both the query and target image.
Therefore, we propose a technique to automatically augment some existing datasets with textual answers and comments to the multimodal questions and image-only answers.
In this work, we focus on 2 different domains, images of real-world object categories with \cirr~\cite{Liu_2021_ICCV} and Wikipedia concepts with \wikimm~\cite{burns2023suite}, but the approach itself generalizes to other domains.
The proposed approach to obtain a tuple of a query image, a textual question, a target image and a textual answer for \tasknameshort is composed by three parts: 
1) pair related images; 
2) generate questions and answers related to those images; and 
3) manual annotation to create a golden set for evaluation.
We couple images that belong to similar concepts, which definition depends on the task.
For \cirr, image pairs are already present in the original set, as images were grouped together based on visual similarity. %
For \wikimm, we pair images belonging to the same Wikipedia article, because that denotes a semantic link.
We subsample a set of 50K samples for training and 2K for testing.

In the second step, we aim at generating questions and answers that can relate the two images, with the goal of simulating the behavior of a user that asks a question about an uploaded image, and obtain an answer with text and the target image supporting the text.
We automatically generate these questions and answers by prompting using Anthropic Claude 3.5 Sonnet.
For \cirr, the question is the textual modification instruction that is already included in the original set, so we only generate the textual answer.
For \wikimm, we generate a question-answer pair that is related to the query and target images, their captions and the Wikipedia article.
This task is more challenging and results in noisier data.
We name these datasets \cirrcomment and \wikicomment, respectively, and show examples\footnote{The Supplementary Material shows more examples of the datasets and the prompts used to generate them.} in Fig.~\ref{fig:qualitative_commenting}.

Finally, we create golden sets for evaluation purposes by asking annotators to validate the plausibility of the question and the quality of the generated text, manually adjusting them if necessary.
We performed human auditing of the generated comments on the validation set of \cirrcomment and found that $\sim\!97\%$ of the audited comments were marked as high quality.
For the test set of \wikicomment, we conducted a manual review by filtering out image pairs that are not related to each other, questions that are irrelevant and comments that are not aligned with the question.
In addition, annotators performed the necessary edits of the question and answer pairs to improve their quality.
This results in a golden set of 695 out of 1768 ($\sim\!39\%$), showing the importance of having an annotation process for evaluation, that is often not considered in previous datasets. 

\section{Experiments}
\label{sec:experiments}

We evaluate the multimodal retrieval and commenting capabilities of our methods with respect to the state of the art in Sec.~\ref{ssec:exp_retrieval} and Sec.~\ref{ssec:exp_commenting}, respectively.

\begin{table*}[t]
    \centering
    \resizebox{\textwidth}{!}{%
    \begin{tabular}{l c ccc c ccc c ccc c ccc c ccc c}
    \toprule
    & 
    & \multicolumn{3}{c}{\textbf{\fashioniq}~\cite{guo2019fashion}} &
    & \multicolumn{3}{c}{\textbf{\cirr}~\cite{Liu_2021_ICCV}} &
    & \multicolumn{3}{c}{\textbf{OVEN}~\cite{hu2023open}} &
    & \multicolumn{3}{c}{\textbf{InfoSeek}~\cite{chen2023can}} &
    & \multicolumn{3}{c}{\textbf{\wikicomment}} &
    \textbf{Average} \\
    \cmidrule{3-5} \cmidrule{7-9} \cmidrule{11-13} \cmidrule{15-17} \cmidrule{19-21} 
    &
    & \textbf{R@10} & \textbf{R@20} & \textbf{R@50} &
    & \textbf{R@1} & \textbf{R@5} & \textbf{R@10} &
    & \textbf{R@1} & \textbf{R@5} & \textbf{R@10} &
    & \textbf{R@1} & \textbf{R@5} & \textbf{R@10} & 
    & \textbf{R@1} & \textbf{R@5} & \textbf{R@10} & 
    \\
    \midrule
    CLIP (zero-shot)~\cite{radford2021clip} & 
    & $5.50$ & $8.26$ & $13.84$ & 
    & $0.86$ & $12.13$ & $18.30$ & 
    & $10.79$ & $24.12$ & $30.47$ &
    & $8.73$ & $21.79$ & $29.62$ & 
    & $13.24$ & $25.04$ & $30.50$ & 
    $16.88$ 
    \\
    UniIR CLIP$_\text{SF}$~\cite{wei2023uniir} & 
    & $24.59$	& $32.22$	& $\mathbf{43.66}$ & 	
    & $9.50$	& $45.01$	& $58.61$ & 	
    & $49.50$	& $68.86$	& $74.56$ & 	
    & $\mathbf{27.08}$	& $\mathbf{49.04}$	& $\mathbf{58.24}$ & 	
    & $21.73$	& $40.72$	& $46.76$	& 
    $43.34$ 
    \\    
    \modelname (ours) &  
    & $\mathbf{25.20}$	& $\mathbf{32.40}$	& $43.33$	& 
    & $\mathbf{16.45}$	& $\mathbf{47.24}$	& $\mathbf{59.76}$	& 
    & $\mathbf{51.08}$	& $\mathbf{70.22}$	& $\mathbf{75.54}$	& 
    & $26.41$	& $48.01$	& $57.24$   &    
    & $\mathbf{40.14}$   & $\mathbf{57.70}$	& $\mathbf{67.05}$ & 
    $\mathbf{47.85}$ 
    \\   
    \bottomrule
    \end{tabular}
    }
    \caption{\underline{Main results on retrieval}. Retrieval results in terms of recall on different datasets while comparing with state of the art methods.}
    \label{tab:retrieval_main}
\end{table*}

\subsection{Retrieval Results}
\label{ssec:exp_retrieval} 

\paragraph{Implementation Details.} 
We choose InternVL2-4B~\cite{chen2024internvl} as the base, frozen LMM and use it to extract hidden states $\hiddenstate(q)$ corresponding to queries.
Specifically, we take the 3072-dimensional hidden state of the last transformer layer after processing of the last input token before generating the response.
We train the encoder $\psi_\mmmod$ and adapter $\psi_\lmmmod$ using the UniIR framework~\cite{wei2023uniir} so as to align the queries and targets using contrastive loss~\cite{ilharco2021openclip} in two stages.
First, we train the hidden state adapter exclusively to align them to UniIR embeddings, so we can obtain a good initialization.
Second, as depicted in Fig.~\ref{fig:comment_retrieval}, we additionally unlock parameter $\beta$ and the multimodal encoder, initialized from a pre-trained CLIP ViT-L-14 model~\cite{radford2021clip}. We append instructions in the query as proposed by~\cite{wei2023uniir} and train a joint model on all 5 datasets\footnote{Fashion-IQ, CIRR, OVEN, and InfoSeek are a subset of M-BEIR~\cite{wei2023uniir} that has multimodal inputs.}, i.e Fashion-IQ, CIRR, OVEN, InfoSeek and Wiki-CoR. %
For Wiki-CoR, we randomly sample comment or caption as text with 50\%/50\% chance. We train for 40 epochs using learning rate 1e-5, batch size 55, effectively 110 thanks to gradient accumulation of 2, on a single host with 8$\times$48GB NVIDIA L40S GPUs.

\paragraph{Metrics.} 
We use standard \emph{recall@k} metrics to measure the retrieval performance: we compute the neighbourhoods of the query vectors and verify if they contain the target image specified in the ground-truth annotations of the datasets.
We compare with pretrained CLIP ViT-L-14~\cite{radford2021clip} and the state-of-the-art model UniIR CLIP$_\text{SF}$~\cite{wei2023uniir}.

\begin{figure*}
  \centering
   \includegraphics[width=0.9\linewidth]{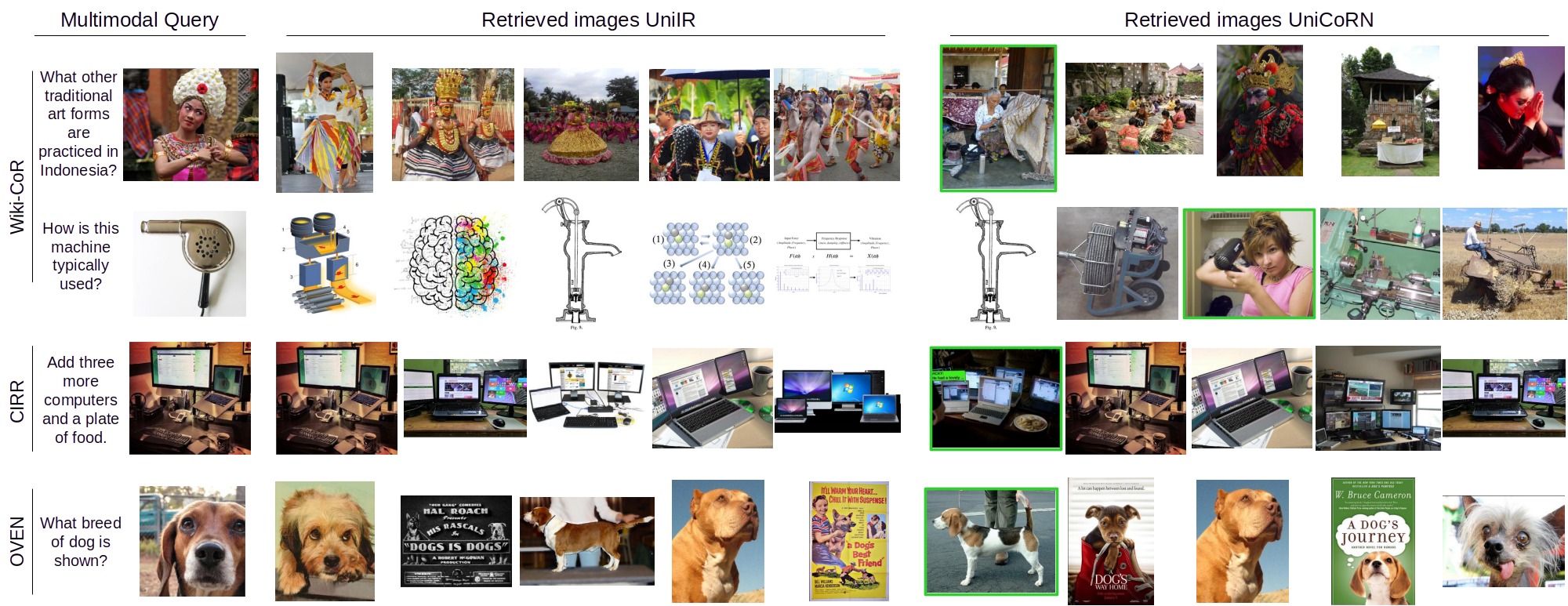}
   \caption{\underline{Qualitative results retrieval.} We show retrieved images for UniIR and \modelname on three datasets. Captions are not displayed because of space limits.}
   \label{fig:qualitative_retrieval}
\end{figure*}

\paragraph{Results.} 
The main results are shown in Table~\ref{tab:retrieval_main}.
First, we observe a low zero-shot performance on all datasets.
This is in line with observations from previous work on composed retrieval~\cite{wei2023uniir,zhang2024magiclens}, which showed that CLIP models cannot easily compose multimodal features to match multimodal answers that are semantically different from the original query.
Second, we see that \modelname is able to leverage the hidden states of the LMM, leading to improved performance, especially for the first ranks (recall@1 and 5), over all datasets but InfoSeek. 
We also show qualitative results of UniIR and \modelname in Fig.~\ref{fig:qualitative_retrieval}. In the first example, UniIR retrieves visually similar traditional clothing and headwear. Instead, \modelname correctly retrieves art forms from Bali and Java (mentioned in the captions), consistent with the query (Indonesia). In the challenging second example, the model needs to infer from the query image that it's a hair dryer and connect the textual ``show in use''. While UniIR fails, \modelname correctly retrieves the ground truth image on rank 3. Also the other examples on CIRR and OVEN show that \modelname is able to combine query and image text in an abstract way to retrieve visually dissimilar correct targets, while UniIR focuses on visual similarity.

\paragraph{Ablation Study.} 
Table~\ref{tab:retrieval_ablation} shows an ablation study, where we train our model without \wikicomment and we remove the LMM adapter.
For this analysis, we study recall@1 because this will be the retrieved image that the LMM will comment (Sec.~\ref{ssec:exp_commenting}).
The inclusion of our adapter (row 1 vs 2) enhances performance across both datasets, leveraging LMM hidden states. This improvement comes at a negligible computational cost during inference since the LMM hidden state is computed already, highlighting our model's efficiency and effectiveness.
When we do not train on \wikicomment data, we observe a big drop in performance, highlighting how distinctive \wikicomment is from other Wikipedia datasets such as OVEN and InfoSeek.
Our model convincingly outperforms the zero-shot baseline (row 1 vs 4), showcasing its robust generalization capabilities and underscores the value of our training approach.

\begin{table}[]
    \centering
    \resizebox{0.43\textwidth}{!}{%
    \centering
    \begin{tabular}{l c c c c c}
    \toprule
    & \textbf{\cirr}~\cite{Liu_2021_ICCV} &
    & \textbf{\wikicomment}&
    \\
    \midrule  
    \modelname
    & $\underline{16.45}$ & 
    & $\mathbf{40.14}$ &
    \\
    ~w/o adapter
    & $16.35$ & 
    & $\underline{39.14}$ &
    \\
    ~w/o \wikicomment
    & $\mathbf{17.29}$ & 
    & $23.02$ &
    \\
    ~w/o training (zero-shot)
    & $0.86$ & 
    & $13.24$ &
    \\ 
    \midrule
    UniIR CLIP$_\text{SF}$~\cite{wei2023uniir}
    & $9.50$ & 
    & $21.73$ &
    \\  
    \bottomrule
    \end{tabular}
    }
    \caption{\underline{Ablation study on retrieval}. Retrieval results in terms of recall@1 on different dataset.} %
    \label{tab:retrieval_ablation}
\end{table}

\newcommand{\metricsize}{\scriptsize}

\begin{table*}[]
    \centering
    \resizebox{\textwidth}{!}{%
\begin{tabular}{l cccccc c cccccc}
\toprule
 & \multicolumn{6}{c}{\textbf{\cirrcomment}} & & \multicolumn{6}{c}{\textbf{\wikicomment}} \\ \cmidrule{2-7} \cmidrule{9-14}
 & \textbf{\metricsize METEOR} & \textbf{\metricsize BLEU} & \textbf{\metricsize ROUGE-1}  & \textbf{\metricsize ROUGE-2} &  \textbf{\metricsize BEM} & \textbf{\metricsize SigLIP-R@1} & %
 & \textbf{\metricsize METEOR} & \textbf{\metricsize BLEU} & \textbf{\metricsize ROUGE-1} & \textbf{\metricsize ROUGE-2} &  \textbf{\metricsize BEM} & \textbf{\metricsize SigLIP-R@1} \\
\midrule
InternVL2~\cite{chen2024internvl} & $0.192$ & $0.013$ & $0.280$ & $0.082$ & $11.9\%$ & $41.4\%$ & & $0.128$ & $0.009$ &  $0.237$ & $0.089$ & $31.8\%$ & $49.1\%$ \\
EchoSight~\cite{yan-xie-2024-echosight} + InternVL2 & $0.311$ & $0.038$ & $0.397$ & $0.119$ & $31.6\%$ & $41.0\%$ & & $0.204$ & $0.018$ & $0.276$ & $0.065$ & $16.2\%$ & $32.1\%$ \\
UniIR CLIP$_\text{SF}$~\cite{wei2023uniir} + InternVL2 & $0.313$ & $0.039$ & $0.399$ & $0.121$ & $32.0\%$ & $42.6\%$ & & $0.212$ & $0.025$ & $0.293$ & $0.078$ & $21.4\%$ & $41.2\%$ \\
\modelname & $\mathbf{0.444}$ & $\mathbf{0.115}$ & $\mathbf{0.516}$ & $\mathbf{0.280}$ & $\mathbf{50.4\%}$ & $\mathbf{45.1\%}$ & & $\mathbf{0.379}$ & $\mathbf{0.128}$ & $\mathbf{0.446}$ & $\mathbf{0.196}$ & $\mathbf{39.3\%}$ & $\mathbf{66.3\%}$ \\
\midrule
\modelname retriever + RAG & $0.318$ & $0.041$ & $0.400$ & $0.123$ & $33.7\%$ & $43.3\%$ & & $0.236$ & $0.037$ & $0.323$ & $0.098$ & $28.7\%$ & $55.2\%$ \\
\modelname w/o trained adapter & $0.317$ & $0.029$ & $0.346$ & $0.106$ & $28.8\%$ & $33.7\%$ & & $0.264$ & $0.033$ & $0.285$ & $0.089$ & $23.7\%$ & $52.4\%$ \\
\modelname w/ shared adapter & $0.407$ & $0.060$ & $0.392$ & $0.181$ & $55.9\%$ & $42.7\%$ & & $0.374$ & $0.127$ & $0.447$ & $0.195$ & $39.3\%$ & $67.6\%$ \\
\modelname w/ UniIR CLIP$_\text{SF}$ & $0.439$ & $0.113$ & $0.510$ & $0.276$ & $49.8\%$ & $44.0\%$ & & $0.347$ & $0.106$ & $0.411$ & $0.168$ & $34.7\%$ & $55.4\%$ \\
\midrule
\emph{UniCoRN w/ GT target oracle} & $0.462$ & $0.130$ & $0.534$ & $0.297$ & $56.1\%$ & $55.9\%$ & & $0.443$ & $0.176$ & $0.517$ & $0.258$ & $55.8\%$ & $92.1\%$ \\
\bottomrule
\end{tabular}
}
\vspace*{-2mm}
\caption{\underline{Main results and ablation study on commenting}. Commenting results by computing several language-based metrics between the predicted comments and the ground-truth (GT) ones on the \cirrcomment and \wikicomment dataset. (Top) Comparison with the state of the art; (Middle) Ablation study for retrieval and generation in \modelname; (Bottom) Upper-bound using GT target documents as an oracle.}
\label{tab:main_commenting}
\end{table*}

\begin{table}[]
    \centering
    \resizebox{0.48\textwidth}{!}{%
\begin{tabular}{l ccccc}
\toprule
 & \multicolumn{5}{c}{\textbf{\wikicomment}} \\ \cmidrule{2-6}
 & \textbf{\metricsize METEOR} & \textbf{\metricsize BLEU} & \textbf{\metricsize ROUGE-1} & \textbf{\metricsize ROUGE-2} & \textbf{\metricsize SigLIP-R@1} \\
\midrule
\modelname & $\mathbf{0.379}$ &  $\mathbf{0.128}$ & $\mathbf{0.446}$ & $\mathbf{0.196}$ & $\mathbf{66.3\%}$ \\
~w/o Wikipedia desc. & $0.302$ & $0.069$ & $0.386$ & $0.143$ & $55.1\%$ \\
~image-only & $0.295$ & $0.065$ & $0.379$ & $0.137$ & $50.6\%$ \\
~caption-only & $0.244$ & $0.034$ & $0.299$ & $0.090$ & $47.3\%$ \\
\midrule
\modelname retr. + RAG & $0.236$  & $0.037$ & $0.323$ & $0.098$ & $55.2\%$ \\
~w/o Wikipedia desc. & $0.227$  & $0.032$ & $0.321$ & $0.098$ & $51.5\%$ \\
~image-only & $0.198$  & $0.028$ & $0.314$ & $0.094$ & $49.6\%$ \\
~caption-only & $0.258$  & $0.024$ & $0.277$ & $0.079$ & $46.0\%$ \\
\bottomrule
\end{tabular}
}
\vspace*{-2mm}
\caption{\underline{Modality ablation study for entity encoding}. We compare results on \wikicomment using \modelname and RAG when removing images, captions and Wikipedia descriptions when providing the entities to the model.}
\label{tab:commenting_ablation_wiki}
\vspace*{-5mm}
\end{table}

\subsection{Commenting Results}
\label{ssec:exp_commenting} 

\paragraph{Implementation Details.} 
To implement retrieval-aware commenting, we need to modify standard LMM procedures, because we perform retrieval during generation and output images interleaved with text. We do so by extending the generation function of LMMs with the following changes.
First, we specify a \emph{retrieval token}, the prediction of which will trigger the retriever \retriever.
We use the start-of-sentence tag, as our \taskname datasets do not exhibit comments prior to retrieval.
The retriever yields a reference to the target entity, which we process via feature extraction and the Entity Adapter to make the LMM understand the retrieved content.
We convert all previously predicted output tokens to input embeddings and append the embedding of the retrieved entity, then feed this as input to the LMM for predicting the next tokens, until the end-of-sentence token is detected.

In addition, we train our Entity Adapters specifically so that InternVL2-4B processes retrieved content differently than content input by a user.
We therefore fine-tune the Entity Adapters separately on \cirrcomment and \wikicomment.
For \cirrcomment, we have 28,225 valid training conversations, where the retrieved entities are simply the target images.
For \wikicomment, we have 50,126 valid training conversations, where the retrieved entities are Wikipedia images with caption, as well as metadata from the corresponding Wikipedia page (title, description).
We train both adapters using the InternVL2 framework~\cite{chen2024internvl} for two epochs on a single host with 4$\times$L40S GPUs, using a learning rate of $1.0e^{-4}$, and evaluate on their respective evaluation sets as we described in Sec.~\ref{sec:datasets}.

\begin{figure*}
  \centering
   \includegraphics[width=0.95\linewidth]{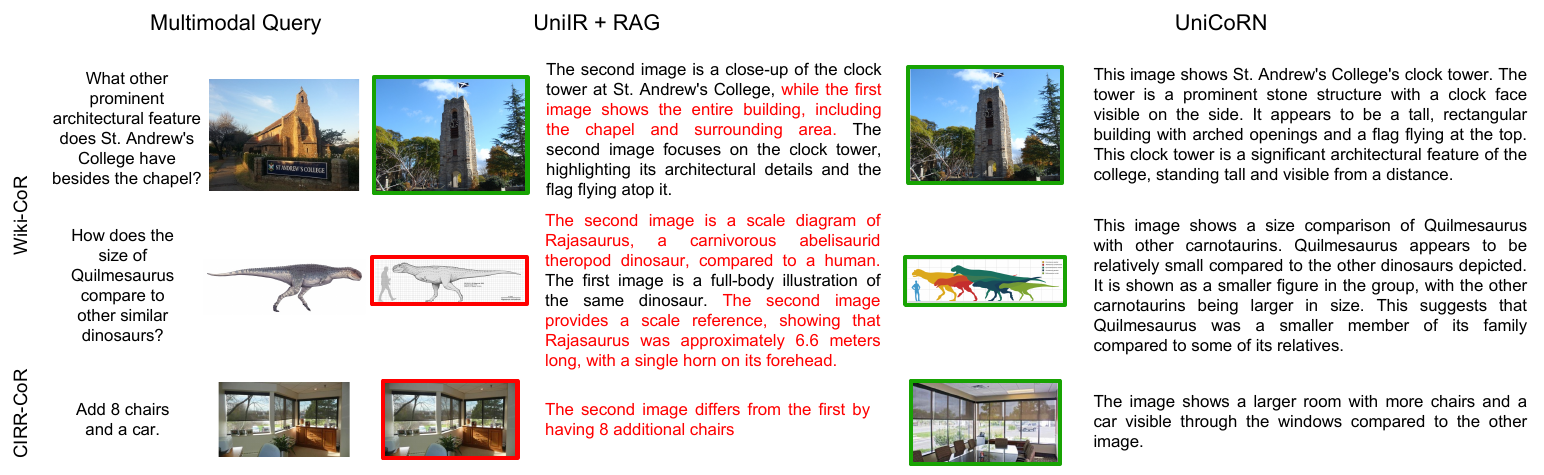}
   \caption{\underline{Qualitative results.} We show retrieved images and comments for UniIR with RAG and \modelname on two different datasets. Comments highlighted in red indicate responses that either do not answer the original question or are not related to the retrieved image. }
   \label{fig:qualitative_commenting}
\end{figure*}

\paragraph{Metrics.}
We evaluate the predicted comments with respect to the corresponding ground-truth (GT) comments in the golden test set using standard metrics for language understanding: \begin{enumerate*}[\roman*)]
\item METEOR~\cite{denkowski:lavie:meteor-wmt:2014}, the harmonic mean of word precision and recall that takes into account stemming and synonymy;
\item BLEU~\cite{10.3115/1073083.1073135}, a metric with high correlation with human judgments of translation quality;
\item ROUGE-1~\cite{lin2004rouge} (resp. ROUGE-2) measuring the overlap of words (resp, bi-grams) between the two comments.
\item BEM~\cite{bem-emnlp2022}, a BERT-based~\cite{bert-2018} measure to evaluate the equivalence of answers to a question.
\end{enumerate*}
We also use a separate vision-language model, namely ViT-SO400M-14-SigLIP-384~\cite{zhai2023sigmoid}, to measure if the predicted comment can be used to retrieve the GT comment with high accuracy, thus measuring if they are specific and distinct from other comments.
For this approach, we use again \emph{recall@1}. %

\paragraph{Results.} 
In the top section of Table~\ref{tab:main_commenting}, we report the results
for \modelname as well as three state-of-the-art LMM approaches that can produce comments based on multimodal inputs:
\begin{enumerate*}[\roman*)]
\item InternVL2~\cite{chen2024internvl}, a state-of-the-art LMM that is prompted to generate a comment for the unseen image based only on the multimodal user query;
\item EchoSight~\cite{yan-xie-2024-echosight}, a state-of-the-art multimodal Retrieval-Augmented Generation (RAG) and re-ranking approach, using InternVL2 as the base LMM for comment generation.
\item a system with the state-of-the-art compositional retriever UniIR~\cite{wei2023uniir} that we use to build a multi-image multi-modal RAG prompt for InternVL2;
\end{enumerate*}
The results show that InternVL2 is able to generate comments that contain relevant information (METEOR 0.192/0.128 on \cirrcomment/\wikicomment resp.).
This entails that the LMM is able to interpret the user query, justifying that its hidden state has indeed the potential to contribute to the retrieval component.
However, the comments are not very similar to the ground-truth: this is clearly due to the absence of retrieval capabilities in InternVL2, which thus has to produce a plausible comment without having access to any information about the content present in the retrieval dataset.
Using EchoSight~\cite{yan-xie-2024-echosight} or UniIR~\cite{wei2023uniir} to retrieve an entity and providing it in the prompt for InternVL2 improves results significantly on all metrics (\eg, METEOR 0.313/212), with a slight edge to the compositional retriever (UniIR) over the reranking one (EchoSight).
This proves that conditioning the generation on actual retrieved images is critical to ensure the relevance of the comments, and explains the broad success of RAG methods.
Then, using \modelname, we further improve to 0.444/0.379 METEOR.
Notably, the improvements are consistent over all metrics on both datasets.
These results show the effectiveness of our approach: without losing any other ability of InternVL2, \modelname is able to optimize its processing of user queries and retrieval results so as to output comments that are significantly better than static prompts built via RAG.

\paragraph{Ablation Study.} 
In the middle section of Table~\ref{tab:main_commenting}, we further analyze the main contributions to the performance of \modelname.
First, when using \modelname's retrieval results in a RAG prompt for InternVL2, we observe only a slight improvement over UniIR+RAG.
Similarly, using \modelname's commenting abilities on UniIR retrieval results performs only slightly worse than our full \modelname.
Thus, the improvements in retrieval are only a part of the contribution to the commenting performance.
In contrast, using the pre-trained input visual encoder from InternVL2 to encode retrieved documents does indeed drop the performance to the level for RAG.
This highlights that our contribution of training dataset-specific entity embeddings makes the biggest contribution in making the base LMM provide more relevant and well-formatted comments.
Training a share entity encoder leads to mixed results, with a slight improvement on BEM and SigLIP at the expense of other metrics.

In Table~\ref{tab:commenting_ablation_wiki}, we further decompose the contribution of the various modalities available in the retrieved documents.
Since \cirrcomment only has target images, we focus on \wikicomment, in which retrieved documents contain an image and its caption on Wikipedia, as well as the title and description of the corresponding page in Wikipedia as metadata.
\modelname, like RAG, uses all information by default in our experiments (to build the Entity encoding, resp. the prompt).
Using the \modelname retriever and the trained adapter, we show the drop of performance that occurs when removing, in turn,
\begin{enumerate*}[\roman*)]
\item the Wikipedia title and description:~$-12.1\%$  METEOR, relatively;
\item all textual inputs (\emph{image-only}):~$$-19.1\%$$;
\item image and Wikipedia metadata (\emph{caption-only}):~$-13.1\%$.
\end{enumerate*}
Since the drops are consistent for all metrics and both comment generation models (RAG and \modelname) based on the same inputs\footnote{Except the caption-only version of RAG for the METEOR score, which does better without the image.}, we can therefore conclude that all three are needed to provide relevant answers. %

Finally, we analyze the qualitative performance of UniIR with RAG and \modelname in Fig.~\ref{fig:qualitative_commenting} with success cases for \modelname.
The first row shows an example where both models retrieve the correct image. 
However, the comments provided by RAG do not answer the question correctly, as indicated by the part of the comment highlighted in red. 
The second row presents an example where the UniIR model was unable to retrieve the correct image, and therefore it could not answer the question correctly. 
This is similar to the third example, where the UniIR model retrieved the wrong query image, yet provided a comment that follows the instructions but does not actually correspond to the retrieved image. 
\section{Conclusions}
\label{sec:conclusion}

This work introduces \modelname, a Unified Commented Retrieval Network that unifies discriminative retrieval and generative commenting. 
We combined the strengths of composed multimodal retrieval methods with generative language capabilities, resulting in a system that surpasses traditional RAG techniques.
Our approach preserves the original capabilities of the base LMM while extending its functionality to perform both retrieval and text generation tasks within a single framework.
We introduced the \tasknameshort task and a corresponding dataset to evaluate these new abilities of LMMs.
Future work could focus on further improving the integration of multimodal information, exploring additional tasks, and investigating the scalability of our approach. %

{
    \small
    \bibliographystyle{ieeenat_fullname}
    \bibliography{biblio}
}

\end{document}